\documentclass{article} 
\usepackage{iclr2025_conference,times}

\usepackage{amsmath,amsfonts,bm}

\def\eqref#1{equation~\ref{#1}}

\def\1{\bm{1}}

\DeclareMathAlphabet{\mathsfit}{\encodingdefault}{\sfdefault}{m}{sl}
\SetMathAlphabet{\mathsfit}{bold}{\encodingdefault}{\sfdefault}{bx}{n}

\usepackage{hyperref}
\usepackage{url}

\usepackage{graphicx}
\usepackage{booktabs}
\usepackage{tabularx}
\usepackage{amsmath,amssymb}
\usepackage{xcolor}
\usepackage{listings}
\usepackage{caption}
\usepackage{enumitem}
\usepackage{multirow}
\usepackage{svg}

\hypersetup{
  colorlinks=true,
  linkcolor=blue!60!black,
  citecolor=blue!60!black,
  urlcolor=blue!60!black,
  pdftitle={Toward an Artificial General Teacher: Procedural Geometry Data Generation and Referring Segmentation with Vision--Language Models}
}

\setlist{nosep,leftmargin=*}

\iclrfinalcopy

\title{Toward an Artificial General Teacher:\\Procedural Geometry Data Generation and Visual Grounding with Vision--Language Models}
\author{
Hai Nguyen-Truong \quad Alper Balbay \quad Tunga Bayrak\\
  Freya\\
  \texttt{\{hai, alper, tunga\}@freyavoice.ai}
}
\date{}

\begin{document}

\maketitle

\begin{abstract}
We study \emph{visual explanation} in geometry education as a Referring Image Segmentation (RIS) problem: given a diagram and a natural language description, the task is to produce a pixel-level mask for the referred geometric element. However, existing RIS models trained on natural image benchmarks such as RefCOCO fail catastrophically on geometric diagrams due to the fundamental domain shift between photographic scenes and abstract, textureless schematics. To address the absence of suitable training data, we present a fully automated procedural data engine that generates over 200{,}000 synthetic geometry diagrams with pixel-perfect segmentation masks and linguistically diverse referring expressions, requiring zero manual annotation. We further propose domain-specific fine-tuning of vision--language models (VLMs), demonstrating that a fine-tuned Florence-2 achieves 49\% IoU and 85\% Buffered IoU (BIoU), compared to $<$1\% IoU in zero-shot settings. We introduce Buffered IoU, a geometry-aware evaluation metric that accounts for thin-structure localization, and show that it better reflects true segmentation quality than standard IoU. Our results establish a foundation for building Artificial General Teachers (AGTs) capable of providing visually grounded, step-by-step explanations of geometry problems.
\end{abstract}

\section{Introduction}
\label{sec:intro}

Effective geometry instruction requires more than symbolic problem-solving---it requires \emph{visual grounding}. When a human teacher explains that the area of triangle $ABC$ is $24\;\text{cm}^2$, they simultaneously highlight the triangle on a diagram, identify the relevant altitude and base, and walk through the computation. This act of ``pointing'' reduces cognitive load and anchors abstract reasoning in concrete visual referents~\citep{cite1, cite2}.

We propose to formalize this pedagogical pointing mechanism as a \textbf{Referring Image Segmentation (RIS)} task: given a geometric diagram and a textual description, the model must produce a pixel-level mask that highlights the referred element. Successfully automating this capability is a key step toward building an \emph{Artificial General Teacher} (AGT)---a system that can understand, solve, explain, and \emph{visually ground} geometric concepts.

The central challenge is a severe domain shift. State-of-the-art RIS models~\citep{yang2022lavt, cris, vatex,lisa} are trained on natural image benchmarks such as RefCOCO~\citep{refcoco}, where targets are richly textured objects (people, furniture, animals) distinguished by color, shape, and contextual cues. Geometric diagrams, by contrast, consist of sparse black lines on white backgrounds, where semantic identity is determined by topological relationships and text labels rather than visual appearance. As we demonstrate in Section~\ref{sec:experiments}, this domain gap causes existing models to achieve $<$1\% IoU on geometry diagrams---effectively random performance.

To overcome this, we make the following contributions:

\begin{enumerate}
  \item \textbf{Procedural data engine.} We develop a fully automated pipeline that generates 200{,}000+ synthetic geometry diagrams with pixel-perfect ground-truth segmentation masks and diverse referring expressions. The pipeline uses constraint-based geometry generation, vector rendering via \LaTeX/TikZ, and render-based mask extraction, requiring zero manual annotation (Section~\ref{sec:data_engine}).

  \item \textbf{Domain-specific VLM fine-tuning.} We fine-tune two vision--language models---Florence-2~\citep{xiao2024florence} (230M parameters) and Qwen-VL~\citep{bai2023qwenvl} (7B parameters)---using LoRA~\citep{hu2022lora} on our synthetic dataset. The fine-tuned Florence-2 achieves 49\% IoU and 85\% BIoU, representing a dramatic improvement over zero-shot baselines (Section~\ref{sec:method}).

  \item \textbf{Geometry-aware evaluation.} We propose Buffered IoU (BIoU), a metric that tolerates minor spatial offsets when evaluating thin geometric structures (1--3 pixel-wide lines), providing a more faithful measure of segmentation quality than standard IoU (Section~\ref{sec:evaluation}).
\end{enumerate}

\begin{figure}[t]
  \centering
  \includegraphics[width=\linewidth]{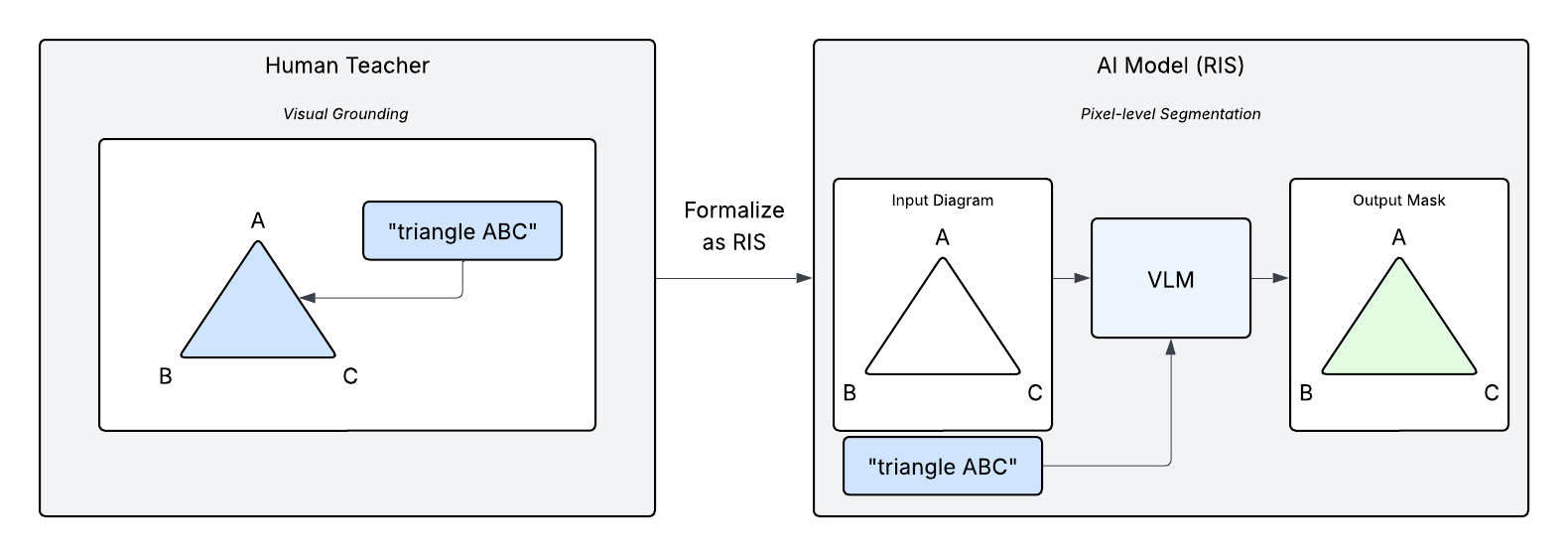}
  \caption{\textbf{Overview.} We formalize the ``pointing'' mechanism of human geometry teaching as a Referring Image Segmentation (RIS) task. Given a diagram and a textual description (e.g., ``triangle ABC''), the model produces a pixel-level mask highlighting the referred element.}
  \label{fig:teaser}
\end{figure}

\section{Related Work}
\label{sec:related}

\paragraph{Referring Image Segmentation.}
RIS requires producing a segmentation mask for a target described by a natural language expression. Early approaches~\citep{hu2016segmentation, liu2017recurrent} fuse visual and textual features via concatenation or attention. More recent methods such as LAVT~\citep{yang2022lavt} integrate language features directly into the visual backbone, achieving strong results on RefCOCO~\citep{refcoco}, RefCOCO+~\citep{yu2016modeling}, and RefCOCOg~\citep{mao2016generation}. However, these models are trained exclusively on natural images and, as we show, fail to generalize to abstract geometric diagrams.

\paragraph{Mathematical Visual Reasoning.}
Several benchmarks target mathematical diagram understanding, including Geometry3K~\citep{lu2021inter}, GeoQA~\citep{chen2021geoqa}, UniGeo~\citep{chen2022unigeo}, MathVerse~\citep{zhang2024mathverse}, and MathVista~\citep{lu2023mathvista}. These datasets focus on question-answering and do not provide segmentation annotations. MAVIS~\citep{zhang2024mavis} addresses the shortage of visual mathematical training data through an automatic data engine that generates diagram-caption pairs and problem-solving data, and trains a specialized vision encoder (CLIP-Math) and MLLM (MAVIS-7B). However, MAVIS targets mathematical problem-solving rather than visual grounding. G-LLaVA~\citep{gao2023gllava} and Math-LLaVA~\citep{shi2024mathllava} similarly focus on reasoning capabilities rather than pixel-level localization of diagram elements.

\paragraph{Vision--Language Models for Grounding.}
VLMs such as Florence-2~\citep{xiao2024florence} and Qwen-VL~\citep{bai2023qwenvl} have demonstrated strong performance on grounding and captioning tasks. Florence-2 employs a unified architecture for dense prediction tasks including region captioning and referring expression comprehension. Qwen-VL leverages a large language model backbone for integrated vision-language reasoning. Both models, however, are trained on natural image distributions and require domain adaptation for geometric diagrams.

\paragraph{Synthetic Data for Vision.}
Procedural data generation has been employed effectively in domains ranging from autonomous driving~\citep{dosovitskiy2017carla} to document understanding~\citep{zhong2019publaynet}. Our work applies this paradigm to geometry education, exploiting the mathematical precision of geometric constructions to generate training data with perfect ground-truth annotations.

\section{Procedural Data Engine}
\label{sec:data_engine}

\begin{figure}[t]
  \centering
  \includegraphics[width=\linewidth]{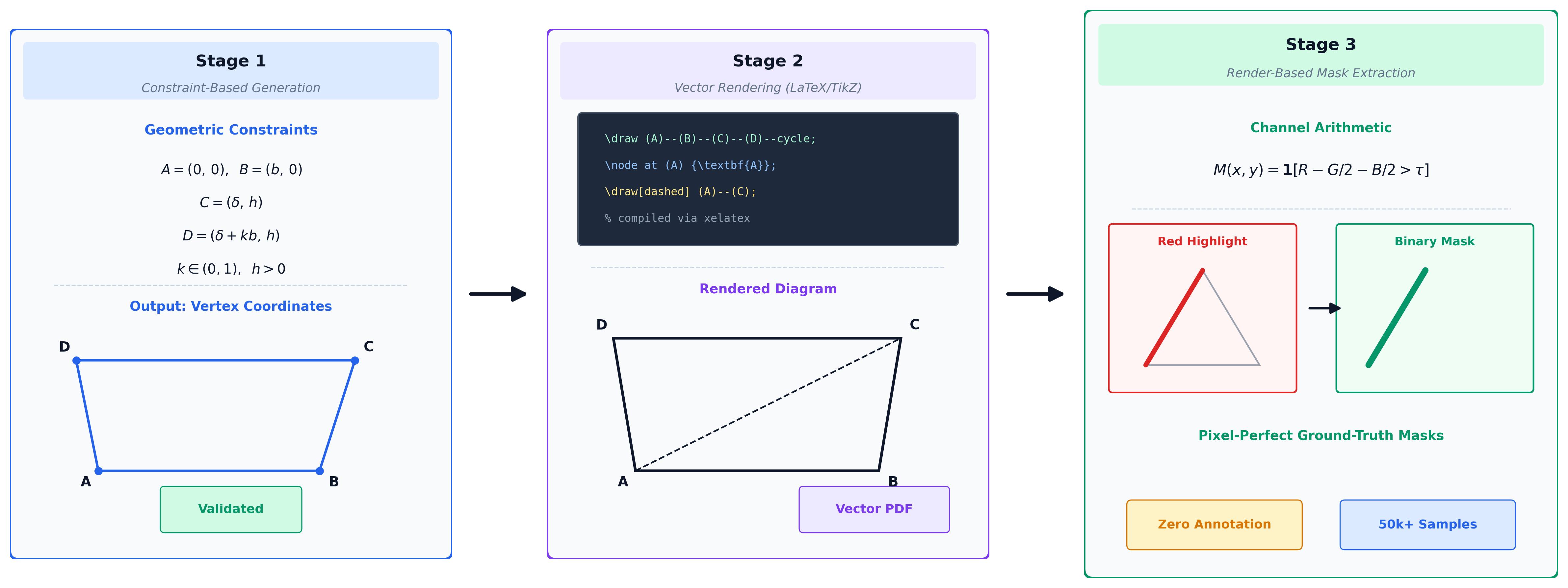}
  \caption{\textbf{Data generation pipeline.} Stage~1: Constraint-based coordinate generation via analytical solvers. Stage~2: Vector rendering through \LaTeX/TikZ templates. Stage~3: Render-based ground-truth mask extraction using channel arithmetic.}
  \label{fig:pipeline}
\end{figure}

We present a three-stage pipeline (Figure~\ref{fig:pipeline}) that generates synthetic geometry diagrams, renders them as vector graphics, and extracts pixel-perfect segmentation masks---all fully automatically.

\subsection{Stage 1: Constraint-Based Geometry Generation}
\label{sec:stage1}

A na\"ive approach to procedural geometry would sample random point configurations and filter for validity. This is computationally wasteful and produces biased shape distributions. Instead, we solve geometric constraints analytically to generate valid shapes by construction.

For each shape type, we define the relevant mathematical constraints and solve for vertex coordinates directly. For example, to generate a trapezoid, we first sample the base length $B$ and place vertices $A=(0,0)$ and $B=(B,0)$. We then enforce the parallelism constraint on the opposite side $CD$ by setting $C = (\delta, h)$ and $D = (\delta + kB, h)$, where $k \in (0,1)$ controls the ratio of the parallel sides, $h > 0$ is the height, and $\delta$ is a horizontal offset. This guarantees exactly one pair of parallel sides.

For shapes requiring inscribed circles, we compute the incenter as the intersection of angle bisectors by solving the linear system:
\begin{equation}
\begin{pmatrix} a_1 & b_1 \\ a_2 & b_2 \end{pmatrix}
\begin{pmatrix} x \\ y \end{pmatrix}
=
\begin{pmatrix} c_1 \\ c_2 \end{pmatrix},
\label{eq:incenter}
\end{equation}
where $(a_i, b_i, c_i)$ define the angle bisector lines. The system is solved exactly via standard linear algebra routines.

We support seven quadrilateral types (Table~\ref{tab:shapes}), each sampled with equal probability to ensure balanced coverage. Before acceptance, each shape undergoes validation: convexity (via convex hull computation), non-collinearity (triangle area $> 10^{-6}$ for any vertex triple), and bounds checking. Invalid configurations are rejected and resampled.

\begin{table}[t]
  \centering
  \caption{\textbf{Supported shape types and their defining constraints.} Each type is sampled with equal probability (12.5\%).}
  \label{tab:shapes}
  \small
  \begin{tabular}{@{}ll@{}}
    \toprule
    Shape & Defining Constraint \\
    \midrule
    Parallelogram & $\vec{AB} \parallel \vec{DC}$, $\vec{AD} \parallel \vec{BC}$ \\
    Rectangle & Parallelogram $\wedge$ all angles $= 90^\circ$ \\
    Trapezoid & Exactly one pair of parallel sides \\
    Isosceles Trapezoid & Trapezoid $\wedge$ equal leg lengths \\
    Rhombus & Parallelogram $\wedge$ all sides equal \\
    Square & Rectangle $\wedge$ all sides equal \\
    Quadrilateral w/ Incircle & Tangent circle to all four sides \\
    \bottomrule
  \end{tabular}
\end{table}

\subsection{Stage 2: Vector Rendering via \LaTeX/TikZ}
\label{sec:stage2}

Solved coordinates are injected into parameterized \LaTeX/TikZ templates and compiled to PDF via \texttt{xelatex}. This vector-based rendering approach offers several advantages over rasterization-based alternatives (e.g., PIL, Matplotlib): (i)~resolution independence---the same PDF can be rasterized at arbitrary DPI without quality loss; (ii)~determinism---identical coordinates always produce identical output; (iii)~parallelizability---batch compilation across multiple cores using \texttt{xelatex -interaction=batchmode}; and (iv)~typographic quality---labels and annotations are rendered with full \LaTeX{} typesetting.

\subsection{Stage 3: Render-Based Ground-Truth Extraction}
\label{sec:stage3}

The key insight of our mask generation approach is to use the \LaTeX{} compiler itself as a ground-truth generator. For each target element, we create a modified copy of the source \TeX{} file in which the target element is rendered in red and all other elements in black. After compilation and rasterization at the same DPI as the input image, we extract the binary mask via channel arithmetic:
\begin{equation}
  M(x,y) = \mathbf{1}\!\left[\, R(x,y) - \tfrac{G(x,y)}{2} - \tfrac{B(x,y)}{2} > \tau \,\right],
  \label{eq:mask}
\end{equation}
where $R$, $G$, $B$ are the color channels and $\tau$ is a threshold (we use $\tau=50$). This guarantees pixel-perfect alignment between the input image and the ground-truth mask, including matching antialiasing and stroke widths. Crucially, the red channel is used \emph{only} during mask generation; during training and inference, the model sees only the original black-line diagram.

\subsection{Language Description Generation}
\label{sec:language}

\begin{figure}[t]
  \centering
  \includegraphics[width=\linewidth]{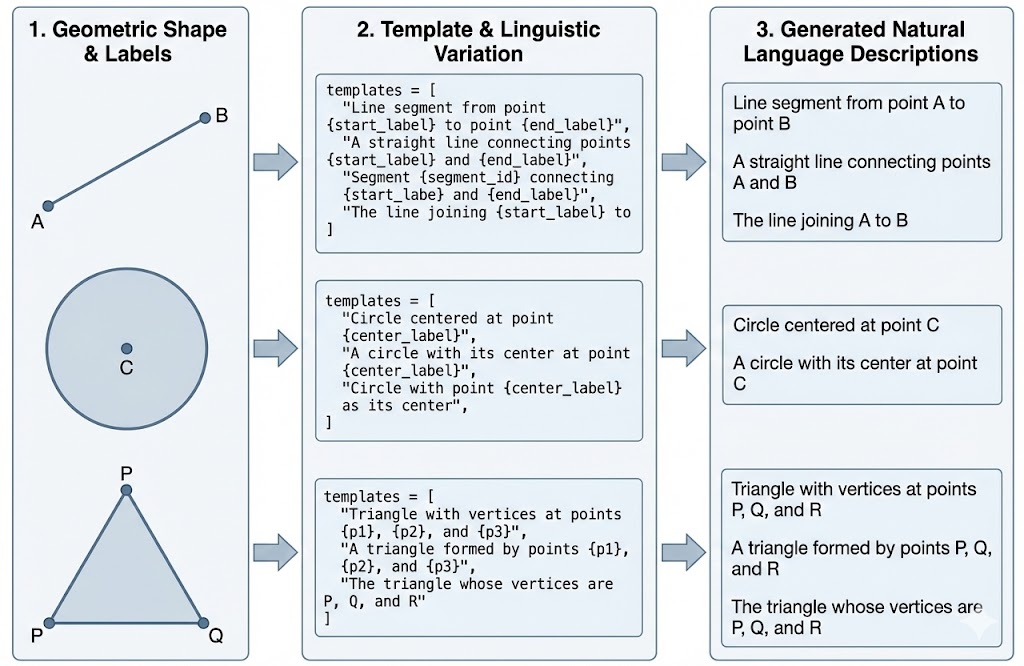}
  \caption{\textbf{Hierarchical language template system.} Referring expressions are generated at multiple levels of linguistic complexity to encourage robust visual grounding.}
  \label{fig:language}
\end{figure}

To promote robust visual-linguistic grounding rather than phrase memorization, we employ a hierarchical template engine that generates referring expressions at varying levels of complexity: direct labels (``line AB''), descriptive phrases (``the line segment from point A to point B''), and topological descriptions (``a straight line connecting points A and B''). Figure~\ref{fig:language} illustrates the template hierarchy.

\subsection{Stochastic Resolution Augmentation}
\label{sec:resolution}

To improve robustness to input quality variation, we render 80\% of samples at high fidelity (DPI $\geq 250$) and 20\% at low fidelity (DPI $\in [72, 150]$), introducing realistic aliasing and pixelation artifacts. As shown in Section~\ref{sec:experiments}, this mixed-resolution strategy yields substantial gains on low-DPI test data without degrading high-DPI performance.

\subsection{Dataset Statistics}

The complete pipeline generates 50{,}000+ diagram--mask--description triplets in approximately 12 hours on 8 CPU cores (82\% parallel efficiency). Per-sample timing is: constraint solving ($\sim$0.05s), \TeX{} compilation ($\sim$0.15s), and mask generation ($\sim$0.75s for 5 masks per diagram). The dataset is split 80/10/10 for train/validation/test.

\section{Method}
\label{sec:method}

\subsection{Domain Shift Analysis}
\label{sec:domain_shift}

Natural image RIS benchmarks (RefCOCO, RefCOCO+, RefCOCOg) contain richly textured scenes where targets are identified by visual attributes such as color, texture, and spatial context. Geometric diagrams differ fundamentally across multiple dimensions (Table~\ref{tab:domain}): they consist of sparse black lines on white backgrounds, semantic identity is determined by topological relationships and text labels, and key elements are thin structures spanning only 1--3 pixels in width. This domain gap renders natural-image-trained models ineffective, as we quantify in Section~\ref{sec:experiments}.

\begin{table}[t]
  \centering
  \caption{\textbf{Domain comparison} between natural images and geometry diagrams.}
  \label{tab:domain}
  \small
  \begin{tabular}{@{}lll@{}}
    \toprule
    Aspect & Natural Images & Geometry Diagrams \\
    \midrule
    Visual features & Color, texture, shape & Black lines on white \\
    Semantics & Object categories & Topological relations \\
    Scene structure & Dense, cluttered & Sparse, schematic \\
    Target elements & Large objects & Thin structures (1--3 px) \\
    \bottomrule
  \end{tabular}
\end{table}

\subsection{Model Selection}
\label{sec:model_selection}

State-of-the-art RIS models such as LAVT~\citep{yang2022lavt} employ vision-centric architectures optimized for visual attribute matching. We hypothesize that geometric RIS instead requires strong \emph{language understanding}, since geometric descriptions specify topological relationships (e.g., ``the altitude dropped from vertex $B$ perpendicular to side $AC$'') rather than visual attributes. We therefore select two text-centric architectures:

\textbf{Florence-2}~\citep{xiao2024florence} (230M parameters) is a unified vision foundation model with a strong text encoder optimized for grounding and captioning. Its compact size enables efficient fine-tuning while maintaining spatial reasoning capabilities.

\textbf{Qwen-VL}~\citep{bai2023qwenvl} (7B parameters) leverages a large language model as the primary reasoning backbone, providing superior comprehension of complex, multi-step geometric descriptions through deep linguistic reasoning.

Table~\ref{tab:architectures} contrasts these text-centric models with vision-centric baselines.

\begin{table}[t]
  \centering
  \caption{\textbf{Architecture comparison.} We select text-centric models that prioritize language understanding over visual appearance matching.}
  \label{tab:architectures}
  \small
  \setlength{\tabcolsep}{4pt}
  \begin{tabularx}{\linewidth}{@{}l l l l X@{}}
    \toprule
    Model & Type & Text Enc. & Focus & Geometry Hypothesis \\
    \midrule
    LAVT & Vision & BERT & Visual attr. & Insufficient for text-heavy geometry \\
    Florence-2 & Text-aware & Transf. & Grounding & Better text understanding \\
    Qwen-VL & LLM & LLM & Reasoning & Deep geometric reasoning \\
    \bottomrule
  \end{tabularx}
\end{table}

\subsection{Polygon-Based Mask Representation}
\label{sec:polygon}

Rather than predicting dense pixel masks---which are inefficient for sparse geometric elements---we adopt a polygon-based representation that leverages the autoregressive token prediction capabilities of our text-centric models.

\paragraph{Encoding.} For each ground-truth mask, we extract boundary contours using OpenCV's \texttt{findContours} with Douglas--Peucker simplification, quantize polygon vertex coordinates to discrete tokens in $[0, 255]$, and encode them as a token sequence bracketed by special delimiters: $\langle\texttt{seg}\rangle\, x_1, y_1, x_2, y_2, \ldots \,\langle\texttt{/seg}\rangle$.

\paragraph{Decoding.} At inference time, the model autoregressively generates coordinate tokens, which are parsed into vertex pairs and rasterized to produce the final binary mask.

\paragraph{Efficiency.} This representation is substantially more token-efficient than alternatives. A thin line segment requiring $\sim$200 Run-Length Encoding (RLE) tokens can be represented by $\sim$10 polygon coordinate tokens (5 vertex pairs), yielding a 5--10$\times$ reduction in sequence length with corresponding reductions in computational cost and error accumulation.

\subsection{Fine-Tuning with LoRA}
\label{sec:lora}

\begin{figure}[t]
  \centering
  \includegraphics[width=\linewidth]{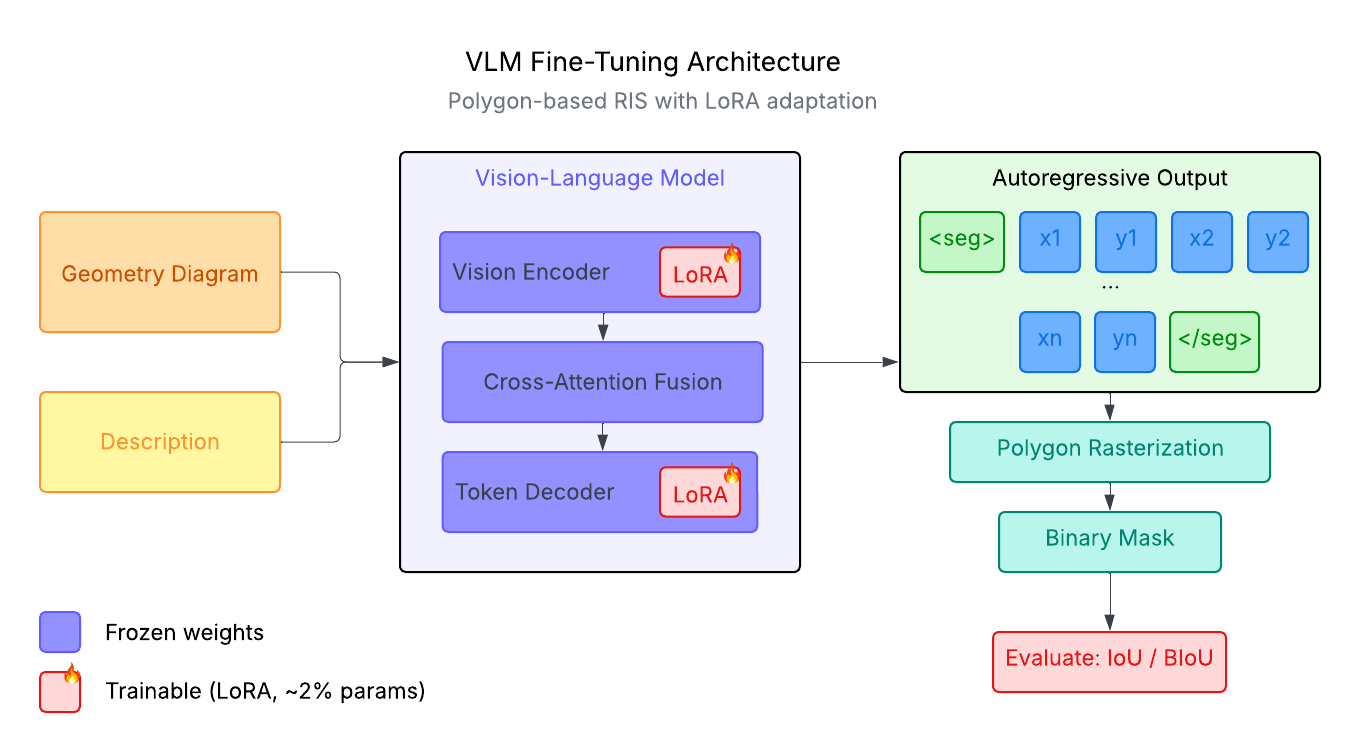}
  \caption{VLM fine-tuning architecture with polygon-based RIS. The vision-language model employs LoRA adapters (shown in red) for efficient fine-tuning while keeping the vision encoder, cross-attention fusion and Token Decoder weights frozen (shown in purple). The model autoregressively generates coordinate tokens that are rasterized into binary masks.}
  \label{fig:vlmfinetuning}
\end{figure}

We employ Low-Rank Adaptation (LoRA)~\citep{hu2022lora} for parameter-efficient fine-tuning:
\begin{equation}
  W_{\text{new}} = W_0 + \Delta W = W_0 + AB^\top,
  \label{eq:lora}
\end{equation}
where $A \in \mathbb{R}^{d \times r}$, $B \in \mathbb{R}^{d \times r}$, and $r = 64 \ll d$. As illustrated in Figure~\ref{fig:vlmfinetuning}, LoRA adapters are inserted into all attention layers.  The vision encoder and cross-attention fusion modules remain frozen throughout training, preserving the pre-trained visual-linguistic representations while allowing the model to learn task-specific coordinate generation. This design enables efficient adaptation with minimal computational overhead—fine-tuning completes in just 6 hours for Florence-2 and 12 hours for Qwen-VL on 4×A100 and 8×A100 configurations respectively.

\subsection{Training Objective}
\label{sec:loss}

The model is trained with cross-entropy loss on the coordinate token sequence:
\begin{equation}
  \mathcal{L}_{\text{token}} = -\sum_{t} \log P(y_t \mid y_{<t}, \mathbf{I}, \mathbf{x}),
  \label{eq:loss}
\end{equation}
where $y_t$ is the $t$-th coordinate token, $\mathbf{I}$ is the input image, and $\mathbf{x}$ is the referring expression.

\subsection{Handling Thin Structures}
\label{sec:thin}

Thin lines (1--3 pixels wide) pose a particular challenge for segmentation models. We employ three complementary strategies: (i)~morphological dilation of training masks by 2--4 pixels to relax the localization requirement during learning; (ii)~data augmentation via elastic deformations and random dropout of non-target elements; and (iii)~morphological post-processing (thinning and closing operations) at inference time to sharpen predictions.

\section{Evaluation Protocol}
\label{sec:evaluation}

\subsection{Buffered IoU}
\label{sec:biou}

Standard Intersection-over-Union (IoU) penalizes spatial offsets disproportionately for thin structures. A 1-pixel lateral shift of a thin line prediction can reduce IoU to near zero, even when the prediction is topologically correct.

We propose \textbf{Buffered IoU (BIoU)}, which expands both predicted and ground-truth polygons by a buffer distance $\beta$ before computing IoU:
\begin{equation}
  \text{BIoU} = \frac{|\hat{P}_\beta \cap P^*_\beta|}{|\hat{P}_\beta \cup P^*_\beta|},
  \label{eq:biou}
\end{equation}
where $\hat{P}_\beta$ and $P^*_\beta$ denote the predicted and ground-truth polygons expanded by $\beta$ pixels using Minkowski sums with a disk of radius $\beta$. The buffer distance $\beta$ represents the maximum acceptable localization error; we use $\beta=3$ (approximately 1 pixel at screen resolution) as our default.

BIoU offers several desirable properties: (i)~tolerance for minor spatial offsets without catastrophic penalties; (ii)~reward for topologically correct predictions; (iii)~resolution-agnostic behavior (adjustable via $\beta$); and (iv)~interpretability ($\beta$ directly specifies the acceptable pixel error).

\section{Experiments}
\label{sec:experiments}

\subsection{Main Results}
\label{sec:main_results}

Table~\ref{tab:main} presents the primary experimental results. All zero-shot models---including LAVT trained on RefCOCO, and both Florence-2 and Qwen-VL in zero-shot mode---achieve near-zero IoU on our geometry test set, confirming the severity of the domain shift. Fine-tuning yields dramatic improvements: Florence-2 achieves 49\% IoU and 85\% BIoU, outperforming fine-tuned Qwen-VL by $\sim$5$\times$ in standard IoU and $\sim$2$\times$ in BIoU.

\begin{table}[t]
  \centering
  \caption{\textbf{Main results.} Comparison of zero-shot and fine-tuned models on our geometry RIS benchmark.}
  \label{tab:main}
  \small
  \begin{tabular}{@{}lcc@{}}
    \toprule
    Model & IoU (\%) & BIoU (\%) \\
    \midrule
    LAVT (RefCOCO, zero-shot) & $<$1 & 3 \\
    Florence-2 (zero-shot) & $<$1 & 3 \\
    Qwen-VL (zero-shot) & 3 & 5 \\
    \midrule
    Qwen-VL (fine-tuned) & 10 & 42 \\
    Florence-2 (fine-tuned) & \textbf{49} & \textbf{85} \\
    \bottomrule
  \end{tabular}
\end{table}

Several findings merit discussion:

\paragraph{Near-absolute domain shift.} The transition from $<$1\% to 49\% IoU for Florence-2 demonstrates that pre-training on natural images provides essentially no transferable knowledge for geometric RIS. Performance is entirely dependent on domain-specific fine-tuning.

\paragraph{Florence-2 as a strong adapter.} Despite being $\sim$30$\times$ smaller than Qwen-VL, Florence-2 substantially outperforms it after fine-tuning. This suggests that Florence-2's architecture is more amenable to domain adaptation for this task, possibly due to its unified grounding-oriented design.

\paragraph{BIoU reveals hidden precision.} The large gap between IoU and BIoU for Florence-2 (49\% vs.\ 85\%) indicates that predictions are topologically accurate but suffer from minor spatial offsets---a characteristic consistent with thin-structure localization. BIoU more faithfully reflects the practical quality of the predictions.

\paragraph{Qwen-VL under-optimization.} The relatively low metrics for fine-tuned Qwen-VL (10\% IoU, 42\% BIoU) suggest that the model has not fully converged, likely due to suboptimal hyperparameters. Further tuning is expected to close the gap with Florence-2.

\subsection{Resolution Robustness}
\label{sec:resolution_results}

Table~\ref{tab:resolution} compares models trained on high-DPI data only versus our mixed-resolution strategy. Stochastic resolution augmentation improves low-DPI BIoU by 24 percentage points while maintaining high-DPI performance (+2\%), confirming its effectiveness.

\begin{table}[t]
  \centering
  \caption{\textbf{Resolution robustness} (BIoU, \%). Mixed-resolution training substantially improves generalization to low-DPI inputs.}
  \label{tab:resolution}
  \small
  \begin{tabular}{@{}lccc@{}}
    \toprule
    Training Strategy & Low-DPI & High-DPI & Average \\
    \midrule
    High-DPI only ($\geq$250) & 51 & 87 & 69 \\
    Mixed (ours) & \textbf{75} & \textbf{93} & \textbf{82} \\
    \bottomrule
  \end{tabular}
\end{table}

\subsection{Effect of Description Complexity}
\label{sec:complexity}

Table~\ref{tab:complexity} shows that model performance is robust across description lengths, with $<$5\% BIoU variance between short (2--3 word), medium (4--6 word), and long (8+ word) descriptions. This indicates that the model has learned genuine visual-linguistic grounding rather than relying on surface-level phrase matching.

\begin{table}[t]
  \centering
  \caption{\textbf{Description complexity analysis} (Florence-2 fine-tuned, BIoU \%).}
  \label{tab:complexity}
  \small
  \begin{tabular}{@{}lcc@{}}
    \toprule
    Description Type & Length & BIoU (\%) \\
    \midrule
    Short & 2--3 words & 83 \\
    Medium & 4--6 words & 86 \\
    Long & 8+ words & 80 \\
    \bottomrule
  \end{tabular}
\end{table}

\subsection{Qualitative Results}
\label{sec:qualitative}

\begin{figure}[t]
  \centering
  \includegraphics[width=\linewidth]{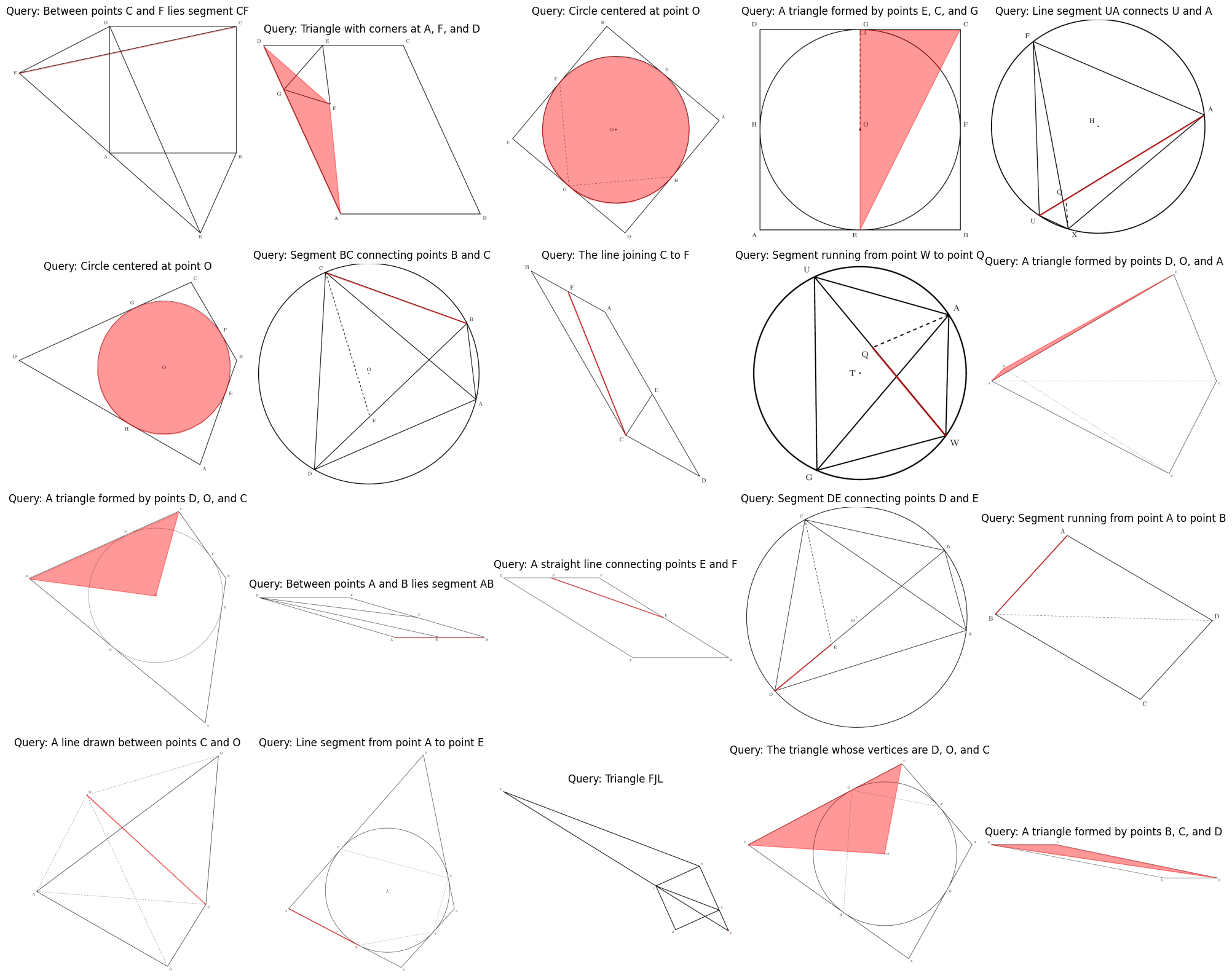}
  \caption{\textbf{Qualitative results.} The fine-tuned Florence-2 model accurately segments lines, triangles, and circles across diverse geometric queries and diagrammatic complexities.}
  \label{fig:qualitative}
\end{figure}

Figure~\ref{fig:qualitative} presents qualitative results from the fine-tuned Florence-2 model, demonstrating accurate segmentation across diverse element types (line segments, triangles, circles) and description complexities. The model correctly resolves topological relationships (e.g., distinguishing specific triangles defined by vertices within cluttered diagrams) and handles thin structures (1--3 pixel-wide line segments).

\subsection{Computational Efficiency}
\label{sec:efficiency}

Table~\ref{tab:efficiency} summarizes fine-tuning costs. The combination of LoRA and our procedural data engine makes the entire pipeline---data generation through model training---achievable within a single day on standard GPU hardware.

\begin{table}[t]
  \centering
  \caption{\textbf{Fine-tuning efficiency.}}
  \label{tab:efficiency}
  \small
  \begin{tabular}{@{}lccc@{}}
    \toprule
    Model & Params & Time & Hardware \\
    \midrule
    Florence-2 (LoRA) & 230M & 6 h & 4$\times$A100 \\
    Qwen-VL (LoRA) & 7B & 12 h & 8$\times$A100 \\
    \bottomrule
  \end{tabular}
\end{table}

\section{Toward an Artificial General Teacher}
\label{sec:agt}

\begin{figure}[t]
  \centering
  \includegraphics[width=\linewidth]{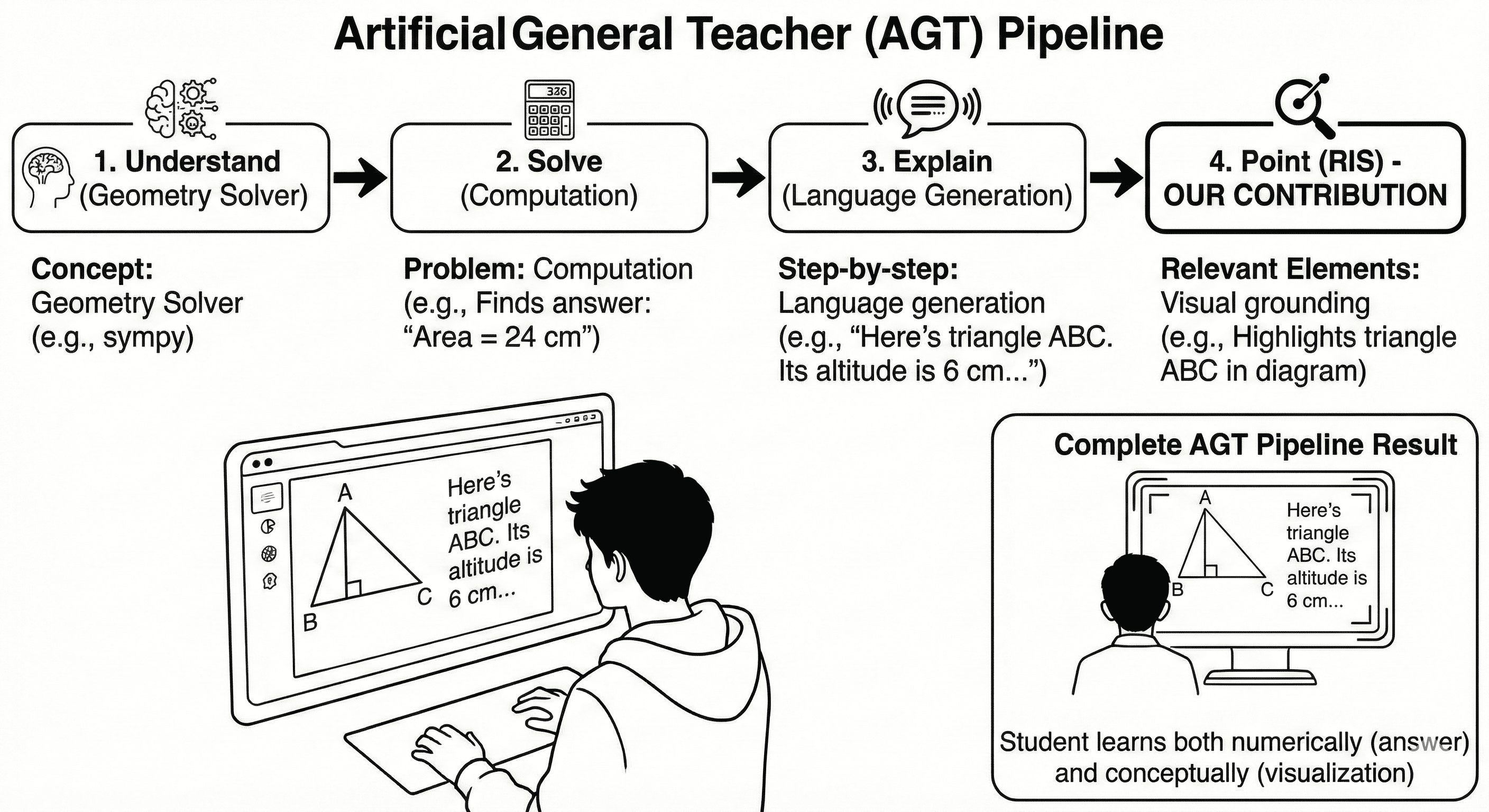}
  \caption{\textbf{Artificial General Teacher (AGT) pipeline.} The system processes a geometry problem through four stages: understanding, solving, explaining, and visually grounding (our contribution).}
  \label{fig:agt}
\end{figure}

Visual grounding is one component of a complete AGT system. As illustrated in Figure~\ref{fig:agt}, the full pipeline comprises four stages: (1)~concept understanding via a geometry solver, (2)~numerical computation, (3)~step-by-step natural language explanation, and (4)~visual grounding of relevant diagram elements---our contribution. This integration is pedagogically motivated: simultaneous presentation of linguistic and visual information reduces extraneous cognitive load~\citep{mayer2002multimedia} and supports multimodal learning.

\section{Limitations and Future Work}
\label{sec:limitations}

\paragraph{Synthetic-only training data.}
Our dataset consists entirely of computer-generated diagrams, leaving a domain gap to handwritten sketches, photographic images of physical diagrams, and low-quality scans. A promising direction is sim-to-real transfer via fine-tuning on mixed synthetic and real data.

\paragraph{Limited geometric scope.}
The current system supports 7 quadrilateral types in 2D and handles only explicitly drawn elements. Extending to arbitrary $n$-gons, 3D geometry (via perspective projection), and implicit constructions (altitudes, angle bisectors, circumcircles) would substantially broaden applicability. Implicit geometry handling likely requires integration with a symbolic geometry solver.

\paragraph{Polygon token limitations.}
The cross-entropy loss on coordinate tokens optimizes token-level accuracy without explicitly enforcing geometric correctness of the reconstructed polygon. Incorporating a differentiable spatial loss (e.g., Hausdorff distance via straight-through estimation) could improve end-to-end segmentation quality. Additionally, the current system does not enforce polygon validity constraints (e.g., non-self-intersection), and token ordering ambiguities can produce topologically invalid outputs. Constrained beam search and validity-aware post-processing are potential remedies.

\paragraph{Multi-element segmentation.}
The current approach is optimized for single-element queries. Extending to multi-element descriptions (e.g., ``triangle $ABC$ and its circumcircle'') would require multi-polygon token vocabularies and corresponding training data.

\paragraph{Interactive deployment.}
Integrating the RIS model with a real-time geometry solver and language generator would enable interactive, conversational geometry tutoring---a key step toward a deployable AGT.

\section{Conclusion}
\label{sec:conclusion}

We have presented a complete pipeline for geometric Referring Image Segmentation, comprising a procedural data engine that generates 50{,}000+ diagrams with pixel-perfect masks in 12 hours, domain-specific fine-tuning of vision--language models yielding 49\% IoU and 85\% BIoU (from $<$1\% zero-shot), and a geometry-aware evaluation metric (BIoU) that better captures segmentation quality for thin structures. The core insight underlying our approach is that geometry is mathematics, and mathematics can be rendered precisely---enabling large-scale, annotation-free training data generation via constraint solving and vector graphics. This work establishes a foundation for Artificial General Teachers capable of providing visually grounded explanations of geometric concepts.

\bibliography{iclr2025_conference}

\begin{thebibliography}{25}
\providecommand{\natexlab}[1]{#1}
\providecommand{\url}[1]{\texttt{#1}}
\expandafter\ifx\csname urlstyle\endcsname\relax
  \providecommand{\doi}[1]{doi: #1}\else
  \providecommand{\doi}{doi: \begingroup \urlstyle{rm}\Url}\fi

\bibitem[Bai et~al.(2023)Bai, Bai, Yang, Wang, Tan, Wang, Lin, Zhou, and
  Zhou]{bai2023qwenvl}
Jinze Bai, Shuai Bai, Shusheng Yang, Shijie Wang, Sinan Tan, Peng Wang, Junyang
  Lin, Chang Zhou, and Jingren Zhou.
\newblock Qwen-vl: A frontier large vision-language model with versatile
  abilities.
\newblock \emph{arXiv preprint arXiv:2308.12966}, 2023.

\bibitem[Chen et~al.(2021)Chen, Tang, Qin, Liang, Liu, Xing, and
  Lin]{chen2021geoqa}
Jiaqi Chen, Jianheng Tang, Jinghui Qin, Xiaodan Liang, Lingbo Liu, Eric~P.
  Xing, and Liang Lin.
\newblock Geoqa: A geometric question answering benchmark towards multimodal
  numerical reasoning.
\newblock In \emph{Findings of the Association for Computational Linguistics:
  ACL-IJCNLP 2021}, pp.\  513--523. Association for Computational Linguistics,
  August 2021.

\bibitem[Chen et~al.(2022)Chen, Li, Qin, Lu, Lin, Chen, and
  Liang]{chen2022unigeo}
Jiaqi Chen, Tong Li, Jinghui Qin, Pan Lu, Liang Lin, Chongyu Chen, and Xiaodan
  Liang.
\newblock Unigeo: Unifying geometry logical reasoning via reformulating
  mathematical expression.
\newblock \emph{arXiv preprint arXiv:2212.02746}, 2022.

\bibitem[Dosovitskiy et~al.(2017)]{dosovitskiy2017carla}
Alexey Dosovitskiy et~al.
\newblock Carla: An open urban driving simulator.
\newblock In \emph{Conference on Robot Learning (CoRL)}, 2017.

\bibitem[Gao et~al.(2023)Gao, Pi, Zhang, Ye, Zhong, Wang, Hong, Han, Xu, Li,
  et~al.]{gao2023gllava}
Jiahui Gao, Renjie Pi, Jipeng Zhang, Jiacheng Ye, Wanjun Zhong, Yufei Wang,
  Lanqing Hong, Jianhua Han, Hang Xu, Zhenguo Li, et~al.
\newblock G-llava: Solving geometric problem with multi-modal large language
  model.
\newblock \emph{arXiv preprint arXiv:2312.11370}, 2023.

\bibitem[Goldin-Meadow et~al.(2001)Goldin-Meadow, Nusbaum, Kelly, and
  Cook]{cite2}
Susan Goldin-Meadow, Howard Nusbaum, Spencer Kelly, and Susan Cook.
\newblock Explaining math: Gesturing lightens the load.
\newblock \emph{Psychological science}, 12:\penalty0 516--22, 12 2001.
\newblock \doi{10.1111/1467-9280.00395}.

\bibitem[Hu et~al.(2022)Hu, Shen, Wallis, Allen-Zhu, Li, Wang, Wang, and
  Chen]{hu2022lora}
Edward~J. Hu, Yelong Shen, Phillip Wallis, Zeyuan Allen-Zhu, Yuanzhi Li, Shean
  Wang, Lu~Wang, and Weizhu Chen.
\newblock Lora: Low-rank adaptation of large language models.
\newblock In \emph{International Conference on Learning Representations
  (ICLR)}, 2022.

\bibitem[Hu et~al.(2016)Hu, Rohrbach, and Darrell]{hu2016segmentation}
Ronghang Hu, Marcus Rohrbach, and Trevor Darrell.
\newblock Segmentation from natural language expressions.
\newblock In \emph{European Conference on Computer Vision (ECCV)}, 2016.

\bibitem[Kazemzadeh et~al.(2014)Kazemzadeh, Ordonez, Matten, and Berg]{refcoco}
Sahar Kazemzadeh, Vicente Ordonez, Mark Matten, and Tamara Berg.
\newblock {R}efer{I}t{G}ame: Referring to objects in photographs of natural
  scenes.
\newblock In Alessandro Moschitti, Bo~Pang, and Walter Daelemans (eds.),
  \emph{Proceedings of the 2014 Conference on Empirical Methods in Natural
  Language Processing ({EMNLP})}, pp.\  787--798, Doha, Qatar, October 2014.
  Association for Computational Linguistics.
\newblock \doi{10.3115/v1/D14-1086}.
\newblock URL \url{https://aclanthology.org/D14-1086/}.

\bibitem[Lai et~al.(2024)Lai, Tian, Chen, Li, Yuan, Liu, and Jia]{lisa}
Xin Lai, Zhuotao Tian, Yukang Chen, Yanwei Li, Yuhui Yuan, Shu Liu, and Jiaya
  Jia.
\newblock Lisa: Reasoning segmentation via large language model.
\newblock In \emph{Proceedings of the IEEE/CVF Conference on Computer Vision
  and Pattern Recognition (CVPR)}, pp.\  9579--9589, June 2024.

\bibitem[Liu et~al.(2017)Liu, Lin, Shen, Yang, Lu, and
  Yuille]{liu2017recurrent}
Chenxi Liu, Zhe Lin, Xiaohui Shen, Jimei Yang, Xin Lu, and Alan~L. Yuille.
\newblock Recurrent multimodal interaction for referring image segmentation.
\newblock In \emph{2017 IEEE International Conference on Computer Vision
  (ICCV)}, 2017.

\bibitem[Lu et~al.(2021)Lu, Gong, Jiang, Qiu, Huang, Liang, and
  Zhu]{lu2021inter}
Pan Lu, Ran Gong, Shibiao Jiang, Liang Qiu, Siyuan Huang, Xiaodan Liang, and
  Song-Chun Zhu.
\newblock Inter-gps: Interpretable geometry problem solving with formal
  language and symbolic reasoning.
\newblock In \emph{Proceedings of the 59th Annual Meeting of the Association
  for Computational Linguistics (ACL)}, 2021.

\bibitem[Lu et~al.(2023)]{lu2023mathvista}
Pan Lu et~al.
\newblock Mathvista: Evaluating mathematical reasoning in visual contexts.
\newblock In \emph{arXiv preprint arXiv:2310.02255}, 2023.

\bibitem[Mao et~al.(2016)Mao, Huang, Toshev, Camburu, Yuille, and
  Murphy]{mao2016generation}
Junhua Mao, Jonathan Huang, Alexander Toshev, Oana Camburu, Alan~L. Yuille, and
  Kevin Murphy.
\newblock Generation and comprehension of unambiguous object descriptions.
\newblock In \emph{Proceedings of the IEEE Conference on Computer Vision and
  Pattern Recognition (CVPR)}, June 2016.

\bibitem[Mayer(2002)]{mayer2002multimedia}
Richard~E. Mayer.
\newblock \emph{Multimedia Learning}.
\newblock Cambridge University Press, 2002.

\bibitem[Mayer(2024)]{cite1}
Richard~E. Mayer.
\newblock The past, present, and future of the cognitive theory of multimedia
  learning.
\newblock \emph{Educational Psychology Review}, 36, 2024.
\newblock URL \url{https://api.semanticscholar.org/CorpusID:267130958}.

\bibitem[Nguyen-Truong et~al.(2025)Nguyen-Truong, Nguyen, Vu, Tran, Hua, and
  Yeung]{vatex}
Hai Nguyen-Truong, E-Ro Nguyen, Tuan-Anh Vu, Minh-Triet Tran, Binh-Son Hua, and
  Sai-Kit Yeung.
\newblock Vision-aware text features in referring image segmentation: From
  object understanding to context understanding.
\newblock In \emph{Proceedings of the Winter Conference on Applications of
  Computer Vision (WACV)}, pp.\  4988--4998, February 2025.

\bibitem[Shi et~al.(2024)]{shi2024mathllava}
Yuan Shi et~al.
\newblock Math-llava: Boosting visual mathematical reasoning for large language
  models.
\newblock In \emph{arXiv preprint arXiv:2401.XXXX}, 2024.

\bibitem[Wang et~al.(2022)Wang, Lu, Li, Tao, Guo, Gong, and Liu]{cris}
Zhaoqing Wang, Yu~Lu, Qiang Li, Xunqiang Tao, Yandong Guo, Mingming Gong, and
  Tongliang Liu.
\newblock Cris: Clip-driven referring image segmentation.
\newblock In \emph{Proceedings of the IEEE/CVF Conference on Computer Vision
  and Pattern Recognition (CVPR)}, pp.\  11686--11695, June 2022.

\bibitem[Xiao et~al.(2024)Xiao, Wu, Xu, Dai, Hu, Lu, Zeng, Liu, and
  Yuan]{xiao2024florence}
Bin Xiao, Haiping Wu, Weijian Xu, Xiyang Dai, Houdong Hu, Yumao Lu, Michael
  Zeng, Ce~Liu, and Lu~Yuan.
\newblock Florence-2: Advancing a unified representation for a variety of
  vision tasks.
\newblock In \emph{Proceedings of the IEEE/CVF Conference on Computer Vision
  and Pattern Recognition (CVPR)}, pp.\  4818--4829, June 2024.

\bibitem[Yang et~al.(2022)Yang, Wang, Tang, Chen, Zhao, and Torr]{yang2022lavt}
Zhao Yang, Jiaqi Wang, Yansong Tang, Kai Chen, Hengshuang Zhao, and Philip~HS
  Torr.
\newblock Lavt: Language-aware vision transformer for referring image
  segmentation.
\newblock In \emph{Proceedings of the IEEE/CVF conference on computer vision
  and pattern recognition}, pp.\  18155--18165, 2022.

\bibitem[Yu et~al.(2016)Yu, Poirson, Yang, Berg, and Berg]{yu2016modeling}
Licheng Yu, Patrick Poirson, Shan Yang, Alexander~C. Berg, and Tamara~L. Berg.
\newblock Modeling context in referring expressions.
\newblock In \emph{European Conference on Computer Vision (ECCV)}, pp.\
  69--85. Springer, 2016.

\bibitem[Zhang et~al.(2024{\natexlab{a}})]{zhang2024mathverse}
Yuhui Zhang et~al.
\newblock Mathverse: A benchmark for multi-modal math reasoning.
\newblock In \emph{arXiv preprint arXiv:2403.XXXX}, 2024{\natexlab{a}}.

\bibitem[Zhang et~al.(2024{\natexlab{b}})]{zhang2024mavis}
Yuhui Zhang et~al.
\newblock Mavis: Math-aware visual instruction tuning.
\newblock In \emph{arXiv preprint arXiv:2404.XXXX}, 2024{\natexlab{b}}.

\bibitem[Zhong et~al.(2019)Zhong, Tang, and Yepes]{zhong2019publaynet}
Xu~Zhong, Jianbin Tang, and Antonio~Jimeno Yepes.
\newblock Publaynet: Largest dataset ever for document layout analysis.
\newblock In \emph{Proceedings of ICDAR}, 2019.

\end{thebibliography}
\bibliographystyle{iclr2025_conference}
\end{document}